\documentclass{fairmeta}

\usepackage{amsmath}
\usepackage{amssymb}
\usepackage{xspace}

\newcommand{\github}{\raisebox{-1.5pt}
{\includegraphics[height=1.05em]{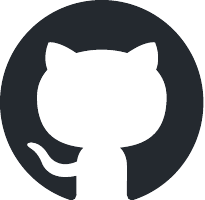}}\xspace}

\title{Progress Reward Modeling for Robotic Learning:\\ A Comprehensive Survey}

\author[1*]{Jianshu Zhang}
\author[1*]{Keliang Wu}
\author[1*]{Haoran Lu}
\author[1]{Anbang Liu}
\author[2]{Ce Zhang}
\author[3]{Weijie Yin}
\author[4]{Chengxuan Qian}
\author[5]{Xiyuan Yang}
\author[1]{Zhenyu Pan}
\author[1]{Guo Ye}
\author[1]{Han Liu}

\small{
\affiliation[1]{Northwestern University}
\affiliation[2]{Carnegie Mellon University}
\affiliation[3]{University of Wisconsin–Madison}
\affiliation[4]{University of California, Santa Barbara}
\affiliation[5]{University of Illinois Urbana-Champaign}
}

{\footnotesize\contribution[*]{Equal Contribution}}

\abstract{

Robotic learning takes place in dynamic environments with large behavior spaces. A terminal success signal only tells the robot whether the task is completed. It does not explain whether the current behavior is making progress, remaining unchanged, or undoing earlier progress. For this reason, recent studies have increasingly explored progress rewards that provide feedback during task execution. However, the current literature lacks a shared framework. Existing methods use different observations, goal specifications, output signals, supervision sources, and evaluation protocols. This makes it difficult to compare them and understand what their results actually validate. In this survey, we provide \textbf{a unified view of progress reward modeling for robotic learning}. We organize the field in three connected steps. We first study the \textit{interface} of a progress model. This defines the problem from the outside by asking what information the model receives and what form of progress signal it produces. We then move inside the model and study the \textit{methods} used to construct this signal. This reveals the different assumptions and mechanisms behind progress estimation and reward generation. Finally, we examine the \textit{data and benchmarks} that support these methods. This shows how progress supervision is obtained and what different evaluations actually measure. Together, these three perspectives connect what a progress model is, how it is built, and how its quality is validated. We further summarize the main limitations of current approaches and discuss future research directions.

    \begin{tabular}{ll}
    
       \github  & \url{https://github.com/sterzhang/Awesome-Progress-Models}\\
    \end{tabular}

}

\begin{document}

\maketitle

\section{Introduction}
\label{sec:introduction}

Robotic learning takes place in dynamic, sequential, and high-dimensional environments~\citep{garmentlab,11128651,generalist2025gen0, Intelligence2026pi07AS, Nvidia2025GR00TNA}.
A robot must interact with objects and the physical world over many steps, and similar actions can have very different effects on task progress depending on the current state and goal~\citep{Intelligence202505AV,Ye2026WorldAM}.
The same task may also be completed through different action sequences~\citep{yuan2026fastwamworldactionmodels,baumli2023visionlanguage}.
These properties create a large behavior space and make it difficult to learn only from final task outcomes.

Most robotic learning systems rely on a terminal success signal\citep{liu2023liberobenchmarkingknowledgetransfer,chen2026robodojo,lu2026magicsimunifiedinfrastructureexecutable}.
Such a signal tells the robot whether the task was eventually completed, but provides little information about the intermediate execution.
It cannot explain whether an action moves the task forward, makes no meaningful change, or undoes earlier progress.
This problem becomes more severe in long-horizon tasks, where successful outcomes are rare and many decisions must be made before receiving any useful feedback~\citep{chen2026rmbenchmemorydependentroboticmanipulation, yong2026generalizabledenserewardlonghorizon}.

\textbf{Progress rewards provide a richer signal by describing how task execution changes over time.}
They can provide dense feedback before the task is completed and improve credit assignment during policy learning. 
They can also help distinguish useful behavior from irrelevant motion, identify stagnation or regression, and detect failures before the final outcome.
Beyond policy learning, progress signals can support online monitoring, trajectory reranking, data filtering, planning, action selection, and failure recovery.
A reliable progress model can therefore serve not only as a reward function, but also as a general evaluator of ongoing robot behavior.

However, estimating progress is more difficult than predicting whether a task has succeeded.
Success detection is usually a binary decision about whether the final goal has been reached.
Progress estimation must instead place the current observation within the evolving execution of a longer task.
The model must infer which parts of the task have been completed, what is currently happening, and how far the execution remains from the goal.
This mapping is also task-conditioned, because the same observation may represent different stages or different amounts of progress under different tasks.
Moreover, the current observation may not contain enough information by itself, since progress can depend on earlier actions, completed subgoals, and whether previous progress has been undone.
Progress is therefore a latent and history-dependent task state rather than a directly observable property of an image or frame.
A progress model must decide what evidence is relevant and how that evidence should be mapped to task advancement.

\begin{figure}[t]
\centering
\includegraphics[width=1\linewidth]{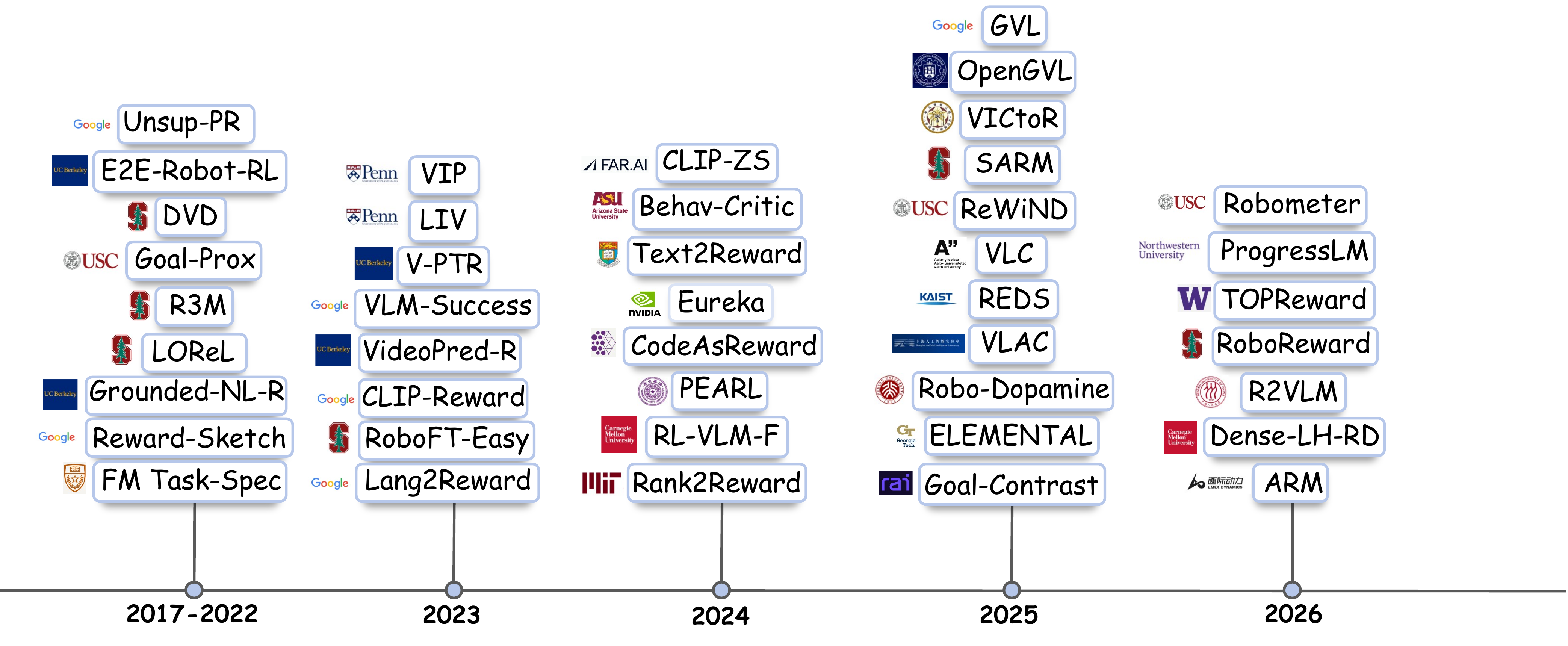}
\caption{Evolution of progress reward modeling for robotic learning.}
\label{fig:works}
\end{figure}

Recent years have seen rapid growth in progress reward modeling, as illustrated in \Cref{fig:works}.
Early studies learned goal proximity and reward representations from demonstrations, temporal ordering, and visual goal structure~\citep{
NEURIPS2021_868b7df9,
ma2023vipuniversalvisualreward,
pmlr-v202-ma23b,
sermanet2017unsupervisedperceptualrewardsimitation}.
Later work explored foundation-model similarity scores and language-conditioned rewards~\citep{
baumli2023visionlanguage,
rocamonde2024visionlanguage,
pmlr-v164-nair22a}, preference-based reward learning~\citep{
wang2024rlvlmf,
pmlr-v235-liu24o,
yang2024rank2rewardlearningshapedreward}, and automatic reward generation with language models~\citep{
xietext2reward,
pmlr-v229-yu23a,
ma2024eureka,
venuto2024code}.
More recent systems use large Vision-Language Models to estimate task progress and completion~\citep{
zhang2026progresslmprogressreasoningvisionlanguage,
liang2026robometerscalinggeneralpurposerobotic,
lee2026roborewardgeneralpurposevisionlanguagereward,
chen2026toprewardtokenprobabilitieshidden}, compare state transitions and model progress changes~\citep{
zhai2025vision,
tan2025robo,
mao2026armadvantagerewardmodeling}, detect task success~\citep{
pmlr-v232-du23b}, and reason over long executions with explicit temporal memory~\citep{
zhang2026recurrentreasoningvisionlanguagemodels}.
This growing body of work shows that progress modeling is becoming an important direction in robotic learning.

Despite this progress, the literature remains fragmented.
Methods described under similar terms often solve different problems.
One method may predict a success probability from a single image, while another estimates a progress percentage from a video, compares two trajectories, or generates an executable reward program.
They also differ in how task goals are specified, how supervision is obtained, and how their outputs are used.
Their evaluations are similarly heterogeneous.
Temporal ordering, scalar prediction error, preference accuracy, success detection, and downstream policy performance validate different properties.
As a result, it is difficult to compare methods directly or determine what their reported results actually demonstrate.

In this survey, we provide \textbf{a unified view of progress reward modeling for robotic learning}.
We organize this diverse field in three connected steps.
First, \Cref{sec:interface} studies the \textit{interface} of progress models.
We treat each model as a task-conditioned black box and examine how it represents the current task state, how it specifies the task goal, and what form of progress-related output it produces.
This view allows methods with different architectures to be compared through a common input--output structure.

Second, \Cref{sec:method} opens this black box and examines \textit{how progress rewards are constructed}.
We organize existing approaches according to the source of the progress signal and the mechanism that converts it into a usable reward.
This perspective reveals the assumptions behind frozen foundation-model scoring, temporal and relative learning, instruction-tuned progress prediction, and programmatic reward construction.

Third, \Cref{sec:data_benchmarks} connects these methods to their \textit{data and evaluation evidence}.
We examine how progress supervision is produced through human annotation, human-guided automation, or fully automated procedures.
We then distinguish benchmarks of progress fidelity from evaluations of robustness, generalization, and downstream utility.
This distinction is important because a reward may improve a policy without faithfully representing progress, while an accurate progress estimator may still be unsuitable for closed-loop control.

\paragraph{Scope and contributions.}
This survey focuses on progress and progress-like reward models for robotic learning.
Rather than treating all reward models as a single category, we study how they define, construct, supervise, and evaluate task advancement.
Our contributions are threefold.
First, we introduce an interface-based view that organizes progress models through task-state representation, goal specification, and output form.
Second, we provide a taxonomy of the main mechanisms used to construct progress rewards and clarify their underlying assumptions.
Third, we connect data-construction pipelines with evaluation protocols and explain what different benchmarks can and cannot validate.
We further summarize the main limitations of current methods and discuss future directions in \Cref{sec:limitations}.


\section{Interface: Input and Output of Progress Models}
\label{sec:interface}

\begin{figure}[t]
    \centering
    \includegraphics[width=1\linewidth]{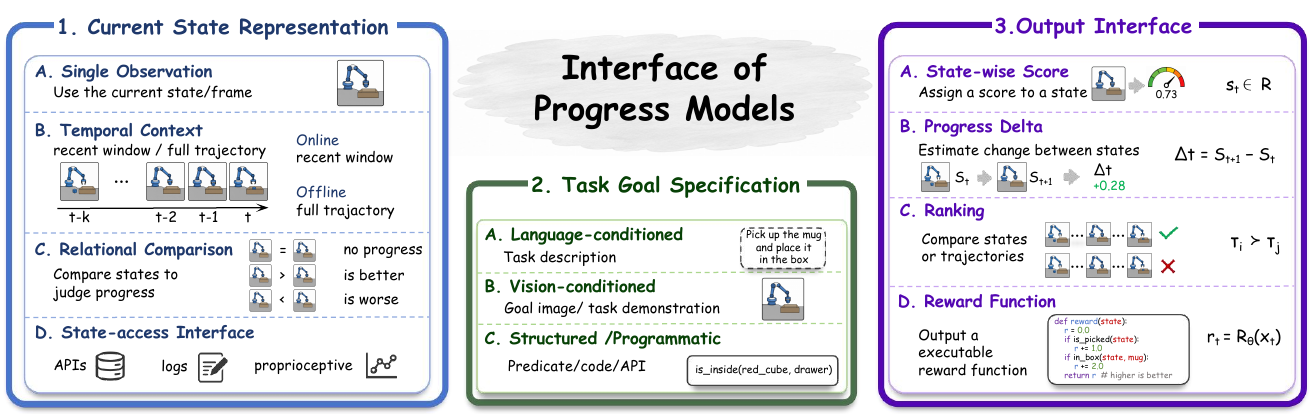}
    \caption{\textbf{Interface of progress models.} We organize existing progress-model interfaces from three perspectives: current task-state representation, task-goal specification, and model output form.}
    \label{fig:interface}
\end{figure}

The interface of a progress model determines what evidence the model can access and what kind of progress or reward signal it can provide. 
As illustrated in \Cref{fig:interface}, we organize the interface around three questions.
For the input interface, the key questions are: (i) \textit{how the current task state is represented} (\cref{subsec: repre}) and (ii) \textit{how the task goal is specified} (\cref{subsec: goal}). 
For the output interface, the key question is: (iii) \textit{what form the model output takes} (\cref{subsec: output}).

\subsection{Input Interface: How Is the Current Task State Represented?}
\label{subsec: repre}
\textbf{Single-observation} inputs are the simplest: the model receives the current observation and a task condition, and directly estimates reward, success, value, or goal proximity. 
This design assumes that the current state contains enough evidence to judge progress. 
It appears in task-completion or perceptual reward models~\citep{singh2019endtoendroboticreinforcementlearning,NEURIPS2021_868b7df9,yang2024rank2rewardlearningshapedreward,sermanet2017unsupervisedperceptualrewardsimitation}, as well as CLIP-based or modern VLM-based reward methods that compare observations with either language or visual goals~\citep{baumli2023visionlanguage,rocamonde2024visionlanguage,pmlr-v162-mahmoudieh22a,pmlr-v168-cui22a,pmlr-v202-ma23b,yang2023robotfinetuningeasypretraining,hung2025victor, zhang2026progresslmprogressreasoningvisionlanguage}. 
\textit{The advantage is low latency and easy online use; the limitation is that visually similar states can have different meanings depending on the execution history.}

\textbf{Temporal context} addresses partially solve this ambiguity by feeding the model a sequence of observations. 
For online progress estimation that the model can directly output while the task is still ongoing, many methods use a trajectory prefix or a short recent window, allowing the model to infer what has already happened before predicting the current progress~\citep{chen2026toprewardtokenprobabilitieshidden,mao2026armadvantagerewardmodeling,zhang2025rewindlanguageguidedrewardsteach,alakuijala2025videolanguagecritictransferablereward,NEURIPS2023_d9042abf,zhang2026recurrentreasoningvisionlanguagemodels,liang2026robometerscalinggeneralpurposerobotic,lee2026roborewardgeneralpurposevisionlanguagereward,chen2025sarmstageawarerewardmodeling,pmlr-v232-du23b,kim2025subtask}. 
Other methods use the full trajectory as input after the task execution to score behavior, rank rollouts, generate preference labels, or analyze temporal progress offline~\citep{guan2024tasksuccess,chen2021learninggeneralizableroboticreward,ma2025vision,budzianowski2025opengvl,pmlr-v235-liu24o}. 
The key tradeoff is clear: \textit{longer context gives stronger temporal evidence, but weakens online usability}.

A third interface estimates progress through \textbf{comparison}. 
Instead of judging whether a single state is good in isolation, the model compares two or more reference states, such as before and after an action, initial and current observations, current and goal states, or two candidate rollouts. 
In this formulation, progress becomes a relational judgment rather than an intrinsic property of one observation. 
Before-after comparison is naturally suited to estimating progress deltas, dense rewards, and preference labels~\citep{zhai2025vision,pmlr-v164-nair22a,ma2023vipuniversalvisualreward,wang2024rlvlmf}. 
Initial-current comparison provides a more global view of how far the execution has moved from the start~\citep{yan2026progressvlaprogressguideddiffusionpolicy,yong2026generalizabledenserewardlonghorizon}, while goal-conditioned comparison grounds the current state or transition against the desired endpoint~\citep{bhateja2023roboticofflinerlinternet,tan2025robo}. 
Comparison-based interfaces are often \textit{more informative and intuitive, but local comparisons can still miss long-horizon dependencies and may suffer from accumulated errors when progress is inferred through a chain of pairwise judgments}.

Finally, some methods use \textbf{state-access interfaces}, such as environment code, simulator APIs, training statistics, or proprioceptive features, to construct rewards~\citep{pmlr-v229-yu23a,xietext2reward,ma2024eureka,venuto2024code,pmlr-v267-chen25at,cabi2020scalingdatadrivenroboticsreward}. 
These interfaces can be highly effective when the relevant task state is explicitly represented and directly accessible. 
However, this assumption is often unrealistic in real-world settings, where progress-relevant variables are usually noisy, incomplete, task-dependent, or unavailable through a clean API. 
Thus, \textit{state-access interfaces are most practical in simulation or instrumented environments, but less suitable for open-ended real-world physical scenes.}

\subsection{Input Interface: How Is the Task Goal Specified?}
\label{subsec: goal}
\textbf{Language-conditioned goals} are used in most progress reward models~\citep{liang2026robometerscalinggeneralpurposerobotic,chen2026toprewardtokenprobabilitieshidden,zhai2025vision,baumli2023visionlanguage,rocamonde2024visionlanguage,xietext2reward,lee2026roborewardgeneralpurposevisionlanguagereward,hung2025victor,guan2024tasksuccess,pmlr-v164-nair22a,pmlr-v162-mahmoudieh22a,mao2026armadvantagerewardmodeling,pmlr-v229-yu23a,pmlr-v202-ma23b,budzianowski2025opengvl,pmlr-v205-nair23a,zhang2025rewindlanguageguidedrewardsteach,chen2025sarmstageawarerewardmodeling,pmlr-v232-du23b,wang2024rlvlmf,alakuijala2025videolanguagecritictransferablereward,yang2023robotfinetuningeasypretraining,kim2025subtask}. 
The main reason is that language goals are \textit{easy to encode}, compatible with existing language or vision-language models, and often \textit{effective for relatively simple tasks}. 
However, language can \textit{underspecify the physical details of progress, especially as tasks become longer, more compositional, and more interaction-dependent}.

\textbf{Vision-conditioned goals} provide a more direct specification of the desired execution process. 
A goal image, goal state, or demonstration trajectory can capture physical details that are difficult to describe with language alone. 
Goal-image and visual-reference methods compare the current observation against a desired visual endpoint~\citep{singh2019endtoendroboticreinforcementlearning,ma2023vipuniversalvisualreward,yang2024rank2rewardlearningshapedreward,bhateja2023roboticofflinerlinternet,tan2025robo,pmlr-v168-cui22a,venuto2024code}, while demonstration-conditioned methods use a full trajectory to specify not only the final state but also the intended procedure~\citep{pmlr-v267-chen25at,chen2021learninggeneralizableroboticreward,NEURIPS2021_868b7df9,pmlr-v235-liu24o,sermanet2017unsupervisedperceptualrewardsimitation,zhang2026progresslmprogressreasoningvisionlanguage}. 
Visual goals therefore \textit{reduce linguistic ambiguity and provide richer physical and dynamic grounding}. 
However, they typically \textit{require manually curated goal images or demonstrations, and can be sensitive to visual variations} such as viewpoint changes, embodiment differences, object appearance, and scene layout.

A third form of goal specification is \textbf{structured or programmatic goals}. 
In simulation or API-rich environments, the task goal can be expressed through predicates, object states, constraints, or generated objective functions. 
This interface supports precise, dense, and compositional reward construction~\citep{xietext2reward,ma2024eureka,pmlr-v229-yu23a,venuto2024code,yong2026generalizabledenserewardlonghorizon}. 
However, it relies on access to explicitly represented state variables, which is often unavailable in open-ended real-world settings. 
Thus, structured or programmatic goals are most effective in simulation or instrumented environments, but less suitable when progress must be inferred from raw visual observations.

\subsection{Output Interface: What Does the Model Predict?}
\label{subsec: output}

The output interface determines \textbf{how a model prediction is interpreted as progress}. 
Different progress models may output scalar scores, transition-level deltas, rankings or preferences, or executable reward functions. 
Although these outputs can all be used as progress signals, they encode progress in different ways: some estimate how desirable a state is, some measure whether a transition moves the task forward, some define progress only through comparison, and some specify a reward computation procedure.

The most common output is a \textbf{state-wise scalar score}. 
Such a score maps a state or clip to a single number, which is then interpreted as task progress, success likelihood, value, goal similarity, or distance-to-goal. 
For explicit progress estimation, the scalar directly represents how far the task has advanced~\citep{liang2026robometerscalinggeneralpurposerobotic,tan2025robo,hung2025victor,chen2025sarmstageawarerewardmodeling,kim2025subtask,zhang2026progresslmprogressreasoningvisionlanguage}. 
For success or completion models, the scalar is interpreted as progress indirectly: higher completion likelihood implies that the current state is closer to task success~\citep{chen2026toprewardtokenprobabilitieshidden,baumli2023visionlanguage,pmlr-v232-du23b,NEURIPS2023_d9042abf,yang2023robotfinetuningeasypretraining,lee2026roborewardgeneralpurposevisionlanguagereward,budzianowski2025opengvl}. 
Similarly, value, similarity, or distance-based methods treat progress as increasing value, increasing similarity to the goal, or decreasing distance from the goal~\citep{rocamonde2024visionlanguage,NEURIPS2021_868b7df9,ma2025vision,ma2023vipuniversalvisualreward,zhang2025rewindlanguageguidedrewardsteach,alakuijala2025videolanguagecritictransferablereward,bhateja2023roboticofflinerlinternet,sermanet2017unsupervisedperceptualrewardsimitation,pmlr-v202-ma23b}. 
Thus, state-wise scores provide \textit{the most direct and reusable output form, but their meaning depends on how the scalar is calibrated and what semantic quantity it is intended to approximate.}

A second output form is \textbf{progress delta}. 
Instead of estimating the absolute progress of a state, the model predicts whether a transition improves or worsens the task state. 
In this case, progress is represented as a local change between two observations, such as before and after an action. 
VLAC directly predicts signed relative progress from before-after observations~\citep{zhai2025vision}, ARM models local advantage-like progress for long-horizon manipulation~\citep{mao2026armadvantagerewardmodeling}, and Robo-Dopamine learns hop values between state pairs~\citep{tan2025robo}. 
Delta outputs are \textit{naturally aligned with reinforcement learning}. 
However, because they describe local improvement rather than global task completion, they usually \textit{need to be accumulated or combined over time to recover total progress}.

A third output form is \textbf{ranking}. 
Here progress is not represented by an absolute score, but by an ordering over states, clips, or trajectories. 
A state or trajectory is considered more advanced if it is preferred over another reference~\citep{pmlr-v168-cui22a,pmlr-v235-liu24o,wang2024rlvlmf,yang2024rank2rewardlearningshapedreward,cabi2020scalingdatadrivenroboticsreward}. 
This output is often \textit{easier to supervise than calibrated reward values} because models can usually compare alternatives more reliably than assign absolute scores. 
However, preference outputs \textit{define progress only relatively and in an indirect way}, and therefore must usually be converted into a scalar reward for later use.

Finally, programmatic methods output \textbf{reward functions} rather than values. 
Here, progress is encoded as an executable computation over available state variables, features, or observations~\citep{xietext2reward,pmlr-v229-yu23a,ma2024eureka,venuto2024code,pmlr-v267-chen25at}. 
This output can be \textit{flexible and interpretable} because it explicitly specifies how progress should be computed. 
However, its reliability depends on whether the generated code, available state variables, and engineered features truly capture the intended task objective.
\section{Methods: How Are Progress Rewards Constructed?}
\label{sec:method}

\begin{figure}[t]
\centering
\includegraphics[width=1\linewidth]{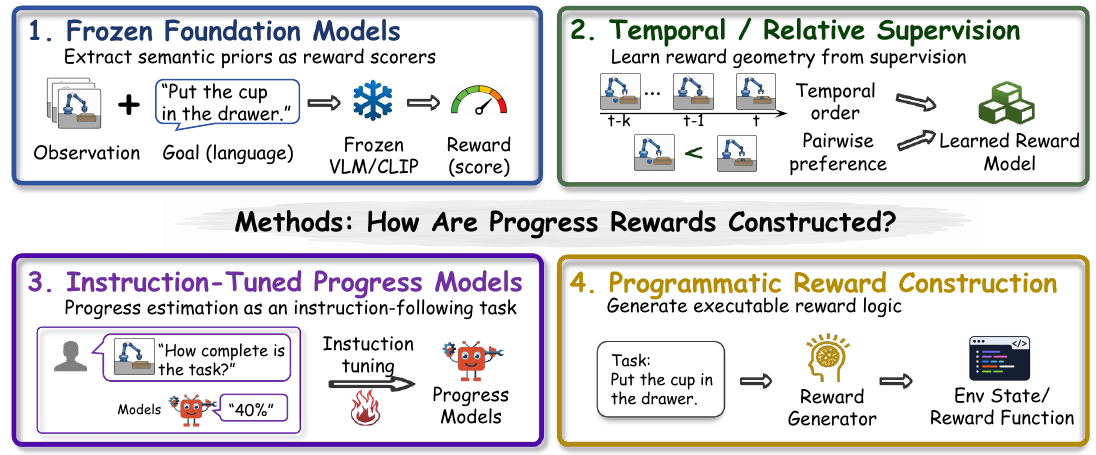}
\caption{\textbf{Construction of progress rewards.} We organize existing methods into four paradigms according to where the progress signal comes from and how it is converted into a usable reward.}
\label{fig:method-overview}
\end{figure}

This section focuses on how progress rewards are constructed.
As illustrated in \Cref{fig:method-overview}, we organize existing methods by the mechanism through which progress signals are obtained and converted into usable rewards.
We identify four main paradigms: frozen foundation-model scoring, learning from temporal or relative supervision, instruction-tuned progress prediction, and programmatic reward construction.

\subsection{Frozen Foundation Models as Semantic Reward Scorers}
\label{subsec: frozen}
The simplest paradigm does not train a reward model at all.
Instead, it directly extracts reward-like signals from pretrained foundation models.
CLIP-style methods compute image-text similarity between the current observation and the language goal, and use the resulting alignment score as reward~\citep{baumli2023visionlanguage,rocamonde2024visionlanguage,pmlr-v162-mahmoudieh22a}.
TOPReward moves this idea from embedding space to token space.
It prompts a frozen VLM with a task-completion question and uses the probability of the ``true'' token as a hidden reward~\citep{chen2026toprewardtokenprobabilitieshidden}.
GVL and OpenGVL further use VLMs to estimate temporal progress or frame ordering from videos~\citep{ma2025vision,budzianowski2025opengvl}.

The main strength of this paradigm is \textbf{zero-shot} usability.
No task-specific reward labels or fine-tuning are required.
However, the resulting score is better understood as a semantic prior than as a calibrated progress reward.
Therefore, these methods often \textit{need normalization, thresholding, differencing, or prompt engineering before their outputs can be used as rewards}.

\subsection{Learning Rewards from Temporal and Relative Supervision}
\label{subsec: relative super}
A second group of methods learns rewards from supervision to determine which state is closer to the goal or which behavior shows more progress.
One common source of supervision is the \textbf{temporal order within successful demonstrations}.
These methods often assume that later states are closer to task completion than earlier states.
Goal-proximity imitation learning uses demonstration time as a proxy for progress and converts changes in goal proximity into rewards~\citep{NEURIPS2021_868b7df9}.
VIP learns a visual representation in which the distance between the current state and the goal reflects goal reachability, allowing this distance to be used as a reward~\citep{ma2023vipuniversalvisualreward}.
Other methods construct rewards from perceptual distance or detected subtask structure~\citep{sermanet2017unsupervisedperceptualrewardsimitation,kim2025subtask}.
A related form of supervision is reward sketching, where users provide coarse reward curves instead of labeling every timestep~\citep{cabi2020scalingdatadrivenroboticsreward}.

Another source of supervision is \textbf{preference or ranking information}.
Instead of assigning an absolute progress value, these methods compare two states, clips, or trajectories and determine which one is better.
RL-VLM-F uses a frozen VLM to generate pairwise preferences and then trains a scalar reward model from these comparisons~\citep{wang2024rlvlmf}.
PEARL transfers preference information across different tasks~\citep{pmlr-v235-liu24o}, while Rank2Reward learns shaped reward functions from rankings of passive videos~\citep{yang2024rank2rewardlearningshapedreward}.
Relative supervision is usually easier to collect than accurate progress scores, but it does not directly provide a usable scalar reward.
The learned comparisons must still be converted into a reward function, and the result may depend on whether the observed preferences can be represented by a single consistent score.

\subsection{Instruction-Tuned Progress Prediction}

A third paradigm \textbf{formulates progress estimation as an explicit instruction-following task for foundation models}.
Rather than relying only on frozen semantic priors or weak temporal structure, these methods specify the desired progress judgment through task-specific prompts and then fine-tune VLMs or video-language models to follow this instruction.

Depending on the formulation, the model may be asked to predict absolute progress, task success, relative improvement, preferences, or reasoning-grounded progress scores.
Training examples therefore pair visual observations or trajectories with explicit progress-related instructions and target responses, enabling the model to acquire progress estimation as a dedicated capability.
RoboReward fine-tunes VLMs with rubric-based prompts to produce structured progress scores over robot rollouts~\citep{lee2026roborewardgeneralpurposevisionlanguagereward}.
Robometer jointly trains progress, success, and preference prediction, allowing a shared model to answer multiple progress-related instructions from dense labels and trajectory comparisons~\citep{liang2026robometerscalinggeneralpurposerobotic}.
Robo-Dopamine estimates hop-based progress from initial, goal, before, and after observations~\citep{tan2025robo}, while VLAC is trained to follow before-after comparison prompts and predict signed progress changes~\citep{zhai2025vision}.
Video-Language Critic learns to map successful and failed video clips to transferable scalar reward judgments~\citep{alakuijala2025videolanguagecritictransferablereward}.
ProgressLM further trains VLMs to generate explicit progress reasoning before predicting progress, using instruction tuning to ground the final score in procedural understanding~\citep{zhang2026progresslmprogressreasoningvisionlanguage}.

The key idea is that progress prediction is not treated as an implicit by-product of visual representation learning.
Instead, it is explicitly defined in the prompt, supervised in the training targets, and learned as an instruction-following capability of the model.

\subsection{Programmatic Reward Construction}

Instead of directly estimating a reward value from observations, some methods represent rewards as executable programs. They translate progress estimation tasks into code, predicates, feature functions, or structured reward logic that can be evaluated during policy learning.
Text2Reward and Language2Rewards convert natural-language task descriptions into executable reward functions~\citep{xietext2reward,pmlr-v229-yu23a}.
Eureka further refines LLM-generated reward programs using feedback from policy training~\citep{ma2024eureka}.
Code as Reward uses VLM reasoning to synthesize executable reward logic from visual task descriptions~\citep{venuto2024code}, while ELEMENTAL combines VLM-generated feature functions with inverse reinforcement learning to construct task rewards~\citep{pmlr-v267-chen25at}.
These methods are particularly effective when a task can be decomposed into explicit subgoals, predicates, geometric relations, or environment states exposed through APIs.
Their flexibility also makes the resulting reward functions relatively interpretable and easy to modify.
However, they shift a substantial part of the challenge from reward prediction to task decomposition, feature design, and reliable state access.
When the available symbolic variables or APIs fail to capture important aspects of the task, the generated reward may be precise with respect to its implementation yet still misaligned with the intended behavior.

\section{Data and Benchmarks}
\label{sec:data_benchmarks}

\begin{figure}[t]
\centering
\includegraphics[width=1\linewidth]{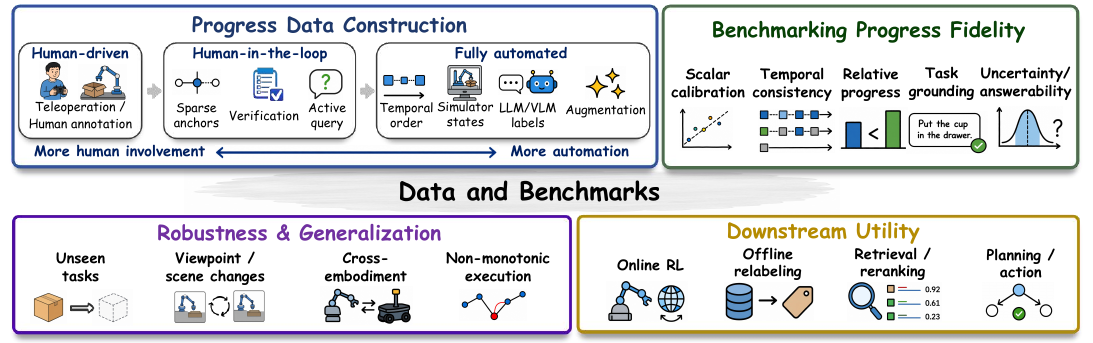}
\caption{\textbf{Data and benchmarks for progress modeling.} We organize progress-data construction by the degree of human involvement and group benchmarks into progress fidelity, robustness and generalization, and downstream utility.}
\label{fig:data-benchmark-overview}
\end{figure}

This section examines how progress supervision is constructed and how progress models are evaluated.
As illustrated in \Cref{fig:data-benchmark-overview}, we first organize progress-data construction by the degree of human involvement.
We then group existing benchmarks according to three evaluation goals: progress fidelity, robustness and generalization, and downstream utility.
This organization connects how progress supervision is obtained with what different evaluation results can actually validate.

\subsection{Data: How Is Progress Supervision Constructed?}
\label{subsec:progress_data}

Progress models require trajectories together with supervision describing task completion, progress change, or relative behavior quality.
Existing works obtain such supervision through direct human judgments, combinations of human input and automated processing, or fully automatic rules and models.
To organize these practices, \textbf{we characterize progress-data construction by the degree of human involvement}.
This perspective reveals a central trade-off: \textit{human supervision provides stronger semantic grounding, whereas automation enables greater scale}.

\paragraph{\textbf{Human-driven construction.}}

Human-driven pipelines rely on people to provide task executions or define progress directly.
Human operators collect robot demonstrations through teleoperation or record human task videos, thereby providing successful or imperfect executions grounded in real behavior~\citep{
chen2026toprewardtokenprobabilitieshidden,
tan2025robo,
zhai2025vision,
ma2025vision,
pmlr-v202-ma23b,
zhang2025rewindlanguageguidedrewardsteach}.
Annotators may assign scalar progress scores~\citep{
lee2026roborewardgeneralpurposevisionlanguagereward,
rocamonde2024visionlanguage}, identify success cutoffs~\citep{
liang2026robometerscalinggeneralpurposerobotic,
pmlr-v232-du23b}, or mark keyframes and subtask boundaries~\citep{
tan2025robo,
chen2025sarmstageawarerewardmodeling,
kim2025subtask}.
Other forms of human supervision include drawing trajectory-level reward sketches~\citep{
cabi2020scalingdatadrivenroboticsreward} and comparing executions or annotating undesirable behavior~\citep{
guan2024tasksuccess,
pmlr-v235-liu24o}.

The main advantage is that humans can capture task semantics that are difficult to infer from time or state variables alone, such as grasp stability, acceptable contact, or implicit behavioral preferences.
However, \textit{dense manual annotation is costly, subjective, and difficult to scale}, especially for long-horizon tasks with multiple valid strategies.

\paragraph{\textbf{Human-in-the-loop construction.}}

Human-in-the-loop pipelines use sparse human input to guide, constrain, or verify automated processing.
A common strategy is to annotate only success frames, keyframes, or stage boundaries and then interpolate dense progress labels between these semantic anchors~\citep{
tan2025robo,
chen2025sarmstageawarerewardmodeling,
pmlr-v232-du23b,zhang2026progresslmprogressreasoningvisionlanguage}.
Models may instead propose stage labels or progress annotations that are subsequently checked and corrected by humans~\citep{
kim2025subtask,
hung2025victor}.
Active-learning methods further reduce cost by requesting human labels only for uncertain or informative states~\citep{
singh2019endtoendroboticreinforcementlearning,
pmlr-v235-liu24o}.

Human involvement may also occur through the design of prompts, task vocabularies, templates, environment abstractions, or reward APIs, after which LLMs or scripts generate annotations and reward specifications at scale~\citep{
pmlr-v229-yu23a,
xietext2reward,
venuto2024code}.
Human inspection or rollout feedback can then be used to revise incorrectly generated labels or rewards~\citep{
ma2024eureka,
pmlr-v267-chen25at}.
Thus, \textit{humans preserve semantic control while automation provides density and scale}, although sparse verification may still fail to detect systematic labeling errors.

\paragraph{\textbf{Fully automated construction.}}

Fully automated pipelines derive progress supervision without instance-level human annotation.
The most common approach uses temporal structure: later frames in successful demonstrations are treated as closer to completion, producing normalized progress labels, goal-proximity targets, or pairwise rankings~\citep{
NEURIPS2021_868b7df9,
ma2023vipuniversalvisualreward,
ma2025vision,
budzianowski2025opengvl,
yang2024rank2rewardlearningshapedreward,
bhateja2023roboticofflinerlinternet}.
Before--after pairs can similarly be labeled according to their temporal distance or relative progress change~\citep{
zhai2025vision,
tan2025robo}.
Such supervision is highly scalable, but \textit{it assumes that task progress is sufficiently correlated with time}.

Structured environments provide another automatic source of labels.
Simulator states, object poses, geometric predicates, and ground-truth rewards can generate success labels, stage indices, continuous progress, or trajectory preferences~\citep{
baumli2023visionlanguage,
pmlr-v168-cui22a,
pmlr-v162-mahmoudieh22a,
pmlr-v202-ma23b}.
Expert policies and motion planners can generate successful trajectories~\citep{
NEURIPS2023_d9042abf}, while noisy or randomized policies produce failed executions~\citep{
alakuijala2025videolanguagecritictransferablereward,
chen2021learninggeneralizableroboticreward}.
These labels are consistent and inexpensive, but \textit{they only capture properties represented by the available states and predicates}.

Foundation models can also serve as automatic annotators.
LLMs and VLMs generate subgoal decompositions, stage boundaries, language variants, progress scores, and pairwise preferences~\citep{
hung2025victor,
mao2026armadvantagerewardmodeling,
yong2026generalizabledenserewardlonghorizon,
wang2024rlvlmf}.
However, \textit{model-generated supervision inherits the biases and errors of the annotating model}.

Automation is also used to broaden behavior coverage.
Successful trajectories can be truncated, reversed, paired with mismatched instructions, or perturbed with action noise to create partial, regressing, and failed examples~\citep{
zhang2025rewindlanguageguidedrewardsteach,
zhai2025vision,
lee2026roborewardgeneralpurposevisionlanguagereward,
liang2026robometerscalinggeneralpurposerobotic,
yang2023robotfinetuningeasypretraining}.
Such augmentation reduces success-only bias, but synthetic failures must remain physically plausible to avoid introducing exploitable artifacts.

\subsection{Benchmarking Progress Fidelity}
\label{subsec:progress_fidelity}

The most direct benchmark question is whether a model output faithfully represents task progress.
However, \textbf{progress fidelity is not a single property}: different reference signals test calibration, temporal consistency, relative ordering, task grounding, or uncertainty.

\textbf{Scalar calibration} evaluates whether predicted scores agree with explicit progress references, such as completion percentages, rubric scores, stage labels, or hop-based progress values~\citep{
lee2026roborewardgeneralpurposevisionlanguagereward,
zhang2026progresslmprogressreasoningvisionlanguage,
chen2025sarmstageawarerewardmodeling,
mao2026armadvantagerewardmodeling,
tan2025robo}.
Typical metrics include regression error and correlation, which can provide the most direct evidence that a model estimates absolute progress.

\textbf{Temporal consistency} evaluates whether predicted scores recover the progression of a trajectory.
Ordering accuracy, rank correlation, and monotonicity-based metrics test whether later states in successful demonstrations tend to receive higher scores than earlier states~\citep{
ma2025vision,
budzianowski2025opengvl,
chen2026toprewardtokenprobabilitieshidden,
zhang2025rewindlanguageguidedrewardsteach,
zhai2025vision,
tan2025robo}.
This evaluation primarily measures chronological consistency rather than absolute calibration.

\textbf{Relative progress} benchmarks evaluate whether the model correctly orders states, clips, or trajectories by task advancement or quality~\citep{
liang2026robometerscalinggeneralpurposerobotic,
yang2024rank2rewardlearningshapedreward,
wang2024rlvlmf}.
They are particularly suitable for preference learning and trajectory selection because pairwise judgments are often easier to obtain than calibrated scalar labels.

\textbf{Task grounding} benchmarks test whether the predicted progress is conditioned on the specified goal rather than on generic motion, scene change, or temporal position.
Counterfactual instructions, mismatched video--language pairs, and contrastive task prompts are commonly used to verify that the same observation receives different judgments under different goals~\citep{
zhang2025rewindlanguageguidedrewardsteach,
zhai2025vision,
ma2025vision}.
This distinction is essential because a smooth and temporally plausible reward curve may still be incorrect if the model ignores the task instruction.

Finally, \textbf{answerability and uncertainty} benchmarks evaluate whether the model can recognize when progress cannot be reliably inferred~\citep{
zhang2026progresslmprogressreasoningvisionlanguage,
zhang2026recurrentreasoningvisionlanguagemodels}.
Progress may be ambiguous because of occlusion, missing history, insufficient viewpoints, or underspecified goals.
A model that always returns a confident score can therefore be unsafe even when its average prediction error is low.
Benchmarking selective prediction, abstention, or confidence calibration is necessary to distinguish reliable progress estimation from forced guessing.

\subsection{Benchmarking Robustness and Generalization}
\label{subsec:robustness_benchmarks}

Beyond matching reference labels, a progress model should preserve its meaning under changes in tasks, observations, embodiments, and execution behavior.
\textbf{Robustness benchmarks test whether the learned notion of progress reflects transferable task structure rather than dataset-specific shortcuts.}

\textbf{Task generalization} evaluates performance on unseen tasks, instructions, objects, or task families~\citep{
zhang2025rewindlanguageguidedrewardsteach,
zhang2026progresslmprogressreasoningvisionlanguage,
pmlr-v235-liu24o,
alakuijala2025videolanguagecritictransferablereward,zhang2026progresslmprogressreasoningvisionlanguage}.
\textbf{Viewpoint and scene robustness} evaluate whether progress estimates remain stable under camera changes, occlusion, distractors, background variation, and differences between demonstration and deployment views~\citep{
zhang2026progresslmprogressreasoningvisionlanguage,
pmlr-v232-du23b}.
Cross-view evaluation  tests whether the model captures task-relevant state rather than superficial visual similarity matching.
\textbf{Cross-embodiment generalization} evaluates whether progress knowledge learned from humans, passive videos, or one robot platform transfers to another embodiment~\citep{
ma2025vision,
alakuijala2025videolanguagecritictransferablereward,
ma2023vipuniversalvisualreward,
pmlr-v202-ma23b,
pmlr-v205-nair23a,
bhateja2023roboticofflinerlinternet,
chen2021learninggeneralizableroboticreward,
sermanet2017unsupervisedperceptualrewardsimitation}.
Such transfer is attractive because human and Internet videos are far more abundant than robot demonstrations.
However, it is challenging because embodiments differ in kinematics, contacts, feasible actions, viewpoints, and execution styles.
The most demanding robustness setting is \textbf{non-monotonic and imperfect execution}.
Real policies may fail, regress, retry an action, revisit an earlier state, or complete subtasks in an unusual order~\citep{
liang2026robometerscalinggeneralpurposerobotic,
lee2026roborewardgeneralpurposevisionlanguagereward,
zhai2025vision,
zhang2025rewindlanguageguidedrewardsteach,
mao2026armadvantagerewardmodeling,
chen2025sarmstageawarerewardmodeling,
hung2025victor,
tan2025robo,
yong2026generalizabledenserewardlonghorizon}.
Benchmarks based only on successful and monotonic demonstrations cannot reveal whether a model distinguishes genuine progress from elapsed time.
Evaluating failures, reversals, plateaus, and recovery behavior is therefore necessary for deployment-relevant progress estimation.

\subsection{Benchmarking Downstream Utility}
\label{subsec:downstream_utility}

Progress models are ultimately intended to support robot learning and decision making.
Accordingly, many benchmarks evaluate whether their outputs improve policy learning, trajectory selection, data curation, or planning.
A reward may be useful without being calibrated, while a well-calibrated estimator may still be too sparse, noisy, or expensive for control.

\textbf{Online policy learning} evaluates whether a progress reward is dense, stable, and sufficiently aligned to support interactive reinforcement learning~\citep{
lee2026roborewardgeneralpurposevisionlanguagereward,
liang2026robometerscalinggeneralpurposerobotic,
zhang2025rewindlanguageguidedrewardsteach,
zhai2025vision,
yang2024rank2rewardlearningshapedreward,
wang2024rlvlmf,
xietext2reward,
ma2024eureka,
venuto2024code}.
Success rate, sample efficiency, and learning stability are common metrics.
However, these results are influenced by exploration, policy architecture, reset design, and environment dynamics.
\textit{Improved policy performance does not by itself demonstrate that the reward is a faithful progress estimator.}

\textbf{Offline learning and reward relabeling} evaluate whether a model can assign useful signals to fixed datasets containing successful, partial, failed, or heterogeneous trajectories~\citep{
liang2026robometerscalinggeneralpurposerobotic,
zhang2025rewindlanguageguidedrewardsteach,
bhateja2023roboticofflinerlinternet,
cabi2020scalingdatadrivenroboticsreward}.
This setting tests whether the reward remains meaningful on behavior generated by different policies and whether it can support offline reinforcement learning, data weighting, or trajectory filtering.
It is especially sensitive to reward calibration outside the distribution of high-quality demonstrations.

\textbf{Retrieval, filtering, and reranking} benchmarks use progress scores to identify useful demonstrations, detect failures, rank candidate rollouts, or select the best result from multiple attempts~\citep{
ma2025vision,
budzianowski2025opengvl,
liang2026robometerscalinggeneralpurposerobotic,
pmlr-v232-du23b,
lee2026roborewardgeneralpurposevisionlanguagereward,
tan2025robo,
alakuijala2025videolanguagecritictransferablereward,
pmlr-v267-chen25at,
guan2024tasksuccess}.
These evaluations primarily test comparative judgment.
They may therefore succeed even when the score lacks an absolute interpretation, and should not be treated as evidence of calibrated progress.

\textbf{Planning and action selection} impose a stronger requirement: the reward landscape must be smooth, informative, and query-efficient enough to evaluate hypothetical states or candidate actions~\citep{
pmlr-v202-ma23b,
ma2023vipuniversalvisualreward,
chen2021learninggeneralizableroboticreward,
pmlr-v164-nair22a,
pmlr-v229-yu23a,
xietext2reward,
venuto2024code}.
A reward that correctly recognizes final success may still be ineffective for planning if it is sparse, discontinuous, visually brittle, or computationally expensive.
Planning benchmarks therefore test whether the progress signal provides useful local guidance, not merely whether it recognizes completed tasks.

\section{Limitations and Future Directions}
\label{sec:limitations}

Despite recent progress, current progress models still face several practical limitations.
In particular, they often lack fine-grained sensitivity, adaptive temporal modeling, efficient online inference, and reliable long-horizon memory.
Addressing these limitations is necessary before progress rewards can be widely used in real-time robotic learning.

\textbf{First, current progress estimation remains too coarse-grained.}
Many models use sparsely sampled frames, short clips, discrete progress bins, or broad task stages~\citep{
liang2026robometerscalinggeneralpurposerobotic,
chen2026toprewardtokenprobabilitieshidden,
tan2025robo,
chen2025sarmstageawarerewardmodeling}.
Such representations can capture major task transitions, but may miss small yet important changes, such as precise alignment, slight object motion, unstable grasping, or early contact failure.
Future models should provide \textit{finer spatial and temporal resolution} and combine visual evidence with proprioception, force, tactile sensing, or other signals that reveal subtle physical progress.

\textbf{Second, progress should not be assumed to change at a fixed rate.}
Many existing methods use temporal order or normalized timestamps as supervision, which encourages progress to increase smoothly with time~\citep{
NEURIPS2021_868b7df9,
ma2023vipuniversalvisualreward,
ma2025vision,
zhang2025rewindlanguageguidedrewardsteach,
zhai2025vision}.
However, real task progress is often uneven.
A major subgoal may cause a sudden increase, while waiting, searching, or careful alignment may produce a long plateau.
The model must also distinguish slow but meaningful execution from behavior that makes no progress.
Future work should develop \textit{temporally adaptive progress models} that adjust the magnitude of progress changes according to task events, execution speed, and stage difficulty, while also representing stagnation and regression.

\textbf{Third, inference latency limits online deployment.}
Large Vision-Language Models often need to process multiple frames, interpret the task instruction, and reason about the current execution before producing a reward~\citep{
alakuijala2025videolanguagecritictransferablereward,
liang2026robometerscalinggeneralpurposerobotic,
zhang2026recurrentreasoningvisionlanguagemodels,
yan2026progressvlaprogressguideddiffusionpolicy}.
This cost may be acceptable for offline evaluation, but it is difficult to use at every control step.
Future systems should explore streaming visual encoders, cached task representations, model distillation, and hierarchical reward querying.
A lightweight model could handle frequent local updates, while a larger model is called only at important task transitions or uncertain states.

\textbf{Finally, long-horizon progress requires stronger and more explicit memory.}
In many tasks, visually similar observations may correspond to different stages or different amounts of progress because their meanings depend on the preceding execution history.
The current observation alone may not reveal which subgoals have already been completed, in what order they occurred, or whether earlier progress has been preserved or undone.
Repeated actions provide a simple example: when a robot must pick and place several objects, the first and second picking actions may look nearly identical, although they represent different completion counts.
A memoryless model may therefore assign similar progress values to states that occupy different positions in the overall task.
Short observation windows provide only limited support when the relevant events lie outside the window~\citep{
zhai2025vision,
liang2026robometerscalinggeneralpurposerobotic,
mao2026armadvantagerewardmodeling,
chen2025sarmstageawarerewardmodeling}.
Future progress models should maintain compact task memory that records completed, pending, and invalidated subgoals, as well as other history-dependent information required to locate the current observation within the full task execution.

\clearpage
\newpage
\bibliographystyle{assets/plainnat}
\bibliography{paper}

@misc{mao2026armadvantagerewardmodeling,
  title={ARM: Advantage Reward Modeling for Long-Horizon Manipulation},
  author={Yiming Mao and Zixi Yu and Weixin Mao and Yinhao Li and Qirui Hu and Zihan Lan and Minzhao Zhu and Hua Chen},
  year={2026},
  eprint={2604.03037},
  archivePrefix={arXiv},
  primaryClass={cs.RO},
  url={https://arxiv.org/abs/2604.03037},
}

@misc{yong2026generalizabledenserewardlonghorizon,
  title={Generalizable Dense Reward for Long-Horizon Robotic Tasks},
  author={Silong Yong and Stephen Sheng and Carl Qi and Xiaojie Wang and Evan Sheehan and Anurag Shivaprasad and Yaqi Xie and Katia Sycara and Yesh Dattatreya},
  year={2026},
  eprint={2604.00055},
  archivePrefix={arXiv},
  primaryClass={cs.RO},
  url={https://arxiv.org/abs/2604.00055},
}

@misc{liang2026robometerscalinggeneralpurposerobotic,
  title={Robometer: Scaling General-Purpose Robotic Reward Models via Trajectory Comparisons},
  author={Anthony Liang and Yigit Korkmaz and Jiahui Zhang and Minyoung Hwang and Abrar Anwar and Sidhant Kaushik and Aditya Shah and Alex S. Huang and Luke Zettlemoyer and Dieter Fox and Yu Xiang and Anqi Li and Andreea Bobu and Abhishek Gupta and Stephen Tu and Erdem Biyik and Jesse Zhang},
  year={2026},
  eprint={2603.02115},
  archivePrefix={arXiv},
  primaryClass={cs.RO},
  url={https://arxiv.org/abs/2603.02115},
}

@misc{chen2026toprewardtokenprobabilitieshidden,
  title={TOPReward: Token Probabilities as Hidden Zero-Shot Rewards for Robotics},
  author={Shirui Chen and Cole Harrison and Ying-Chun Lee and Angela Jin Yang and Zhongzheng Ren and Lillian J. Ratliff and Jiafei Duan and Dieter Fox and Ranjay Krishna},
  year={2026},
  eprint={2602.19313},
  archivePrefix={arXiv},
  primaryClass={cs.RO},
  url={https://arxiv.org/abs/2602.19313},
}

@misc{zhang2026progresslmprogressreasoningvisionlanguage,
  title={PROGRESSLM: Towards Progress Reasoning in Vision-Language Models},
  author={Jianshu Zhang and Chengxuan Qian and Haosen Sun and Haoran Lu and Dingcheng Wang and Letian Xue and Han Liu},
  year={2026},
  eprint={2601.15224},
  archivePrefix={arXiv},
  primaryClass={cs.CV},
  url={https://arxiv.org/abs/2601.15224},
}

@misc{lee2026roborewardgeneralpurposevisionlanguagereward,
  title={RoboReward: General-Purpose Vision-Language Reward Models for Robotics},
  author={Tony Lee and Andrew Wagenmaker and Karl Pertsch and Percy Liang and Sergey Levine and Chelsea Finn},
  year={2026},
  eprint={2601.00675},
  archivePrefix={arXiv},
  primaryClass={cs.RO},
  url={https://arxiv.org/abs/2601.00675},
}

@article{tan2025robo,
  title={Robo-Dopamine: General Process Reward Modeling for High-Precision Robotic Manipulation},
  author={Tan, Huajie and Chen, Sixiang and Xu, Yijie and Wang, Zixiao and Ji, Yuheng and Chi, Cheng and Lyu, Yaoxu and Zhao, Zhongxia and Chen, Xiansheng and Co, Peterson and Xie, Shaoxuan and Yao, Guocai and Wang, Pengwei and Wang, Zhongyuan and Zhang, Shanghang},
  journal={arXiv preprint arXiv:2512.23703},
  year={2025}
}

@misc{chen2025sarmstageawarerewardmodeling,
  title={SARM: Stage-Aware Reward Modeling for Long Horizon Robot Manipulation},
  author={Qianzhong Chen and Justin Yu and Mac Schwager and Pieter Abbeel and Yide Shentu and Philipp Wu},
  year={2025},
  eprint={2509.25358},
  archivePrefix={arXiv},
  primaryClass={cs.RO},
  url={https://arxiv.org/abs/2509.25358},
}

@article{budzianowski2025opengvl,
  title={OpenGVL -- Benchmarking Visual Temporal Progress for Data Curation},
  author={Budzianowski, Pawe{\l} and Wi{\'s}nios, Emilia and Tyrolski, Micha{\l} and G{\'o}ral, Gracjan and Kulakov, Igor and Petrenko, Viktor and Walas, Krzysztof},
  journal={arXiv preprint arXiv:2509.17321},
  year={2025}
}

@article{zhai2025vision,
  title={A Vision-Language-Action-Critic Model for Robotic Real-World Reinforcement Learning},
  author={Zhai, Shaopeng and Zhang, Qi and Zhang, Tianyi and Huang, Fuxian and Zhang, Haoran and Zhou, Ming and Zhang, Shengzhe and Liu, Litao and Lin, Sixu and Pang, Jiangmiao},
  journal={arXiv preprint arXiv:2509.15937},
  year={2025}
}

@misc{zhang2025rewindlanguageguidedrewardsteach,
  title={ReWiND: Language-Guided Rewards Teach Robot Policies without New Demonstrations},
  author={Jiahui Zhang and Yusen Luo and Abrar Anwar and Sumedh Anand Sontakke and Joseph J Lim and Jesse Thomason and Erdem Biyik and Jesse Zhang},
  year={2025},
  eprint={2505.10911},
  archivePrefix={arXiv},
  primaryClass={cs.RO},
  url={https://arxiv.org/abs/2505.10911},
}

@inproceedings{kim2025subtask,
  title={Subtask-Aware Visual Reward Learning from Segmented Demonstrations},
  author={Kim, Changyeon and Heo, Minho and Lee, Doohyun and Lee, Honglak and Shin, Jinwoo and Lim, Joseph J. and Lee, Kimin},
  booktitle={The Thirteenth International Conference on Learning Representations},
  year={2025},
  url={https://openreview.net/forum?id=mqKVe6F3Up}
}

@inproceedings{ma2025vision,
  title={Vision Language Models are In-Context Value Learners},
  author={Ma, Yecheng Jason and Hejna, Joey and Wahid, Ayzaan and Fu, Chuyuan and Shah, Dhruv and Liang, Jacky and Xu, Zhuo and Kirmani, Sean and Xu, Peng and Driess, Danny and Tompson, Jonathan and Bastani, Osbert and Jayaraman, Dinesh and Yu, Wenhao and Zhang, Tingnan and Sadigh, Dorsa and Xia, Fei},
  booktitle={The Thirteenth International Conference on Learning Representations},
  year={2025},
  url={https://openreview.net/forum?id=friHAl5ofG}
}

@inproceedings{pmlr-v267-chen25at,
  title={{ELEMENTAL}: Interactive Learning from Demonstrations and Vision-Language Models for Reward Design in Robotics},
  author={Chen, Letian and Moorman, Nina Marie and Gombolay, Matthew Craig},
  booktitle={Proceedings of the 42nd International Conference on Machine Learning},
  pages={8700--8725},
  year={2025},
  volume={267},
  series={Proceedings of Machine Learning Research},
  publisher={PMLR},
  url={https://proceedings.mlr.press/v267/chen25at.html}
}

@misc{alakuijala2025videolanguagecritictransferablereward,
  title={Video-Language Critic: Transferable Reward Functions for Language-Conditioned Robotics},
  author={Minttu Alakuijala and Reginald McLean and Isaac Woungang and Nariman Farsad and Samuel Kaski and Pekka Marttinen and Kai Yuan},
  year={2025},
  eprint={2405.19988},
  archivePrefix={arXiv},
  primaryClass={cs.RO},
  url={https://arxiv.org/abs/2405.19988},
}

@inproceedings{hung2025victor,
  title={VICtoR: Learning Hierarchical Vision-Instruction Correlation Rewards for Long-horizon Manipulation},
  author={Hung, Kuo-Han and Lo, Pang-Chi and Yeh, Jia-Fong and Hsu, Han-Yuan and Chen, Yi-Ting and Hsu, Winston H.},
  booktitle={The Thirteenth International Conference on Learning Representations},
  year={2025},
  url={https://openreview.net/forum?id=UpQLu9bzAR}
}

@misc{yang2024rank2rewardlearningshapedreward,
  title={Rank2Reward: Learning Shaped Reward Functions from Passive Video},
  author={Daniel Yang and Davin Tjia and Jacob Berg and Dima Damen and Pulkit Agrawal and Abhishek Gupta},
  year={2024},
  eprint={2404.14735},
  archivePrefix={arXiv},
  primaryClass={cs.RO},
  url={https://arxiv.org/abs/2404.14735},
}

@inproceedings{wang2024rlvlmf,
  title={RL-VLM-F: Reinforcement Learning from Vision Language Foundation Model Feedback},
  author={Wang, Yufei and Sun, Zhanyi and Zhang, Jesse and Xian, Zhou and Biyik, Erdem and Held, David and Erickson, Zackory},
  booktitle={Proceedings of the 41st International Conference on Machine Learning},
  pages={51484--51501},
  year={2024},
  volume={235},
  series={Proceedings of Machine Learning Research},
  url={https://proceedings.mlr.press/v235/wang24bn.html}
}

@inproceedings{venuto2024code,
  title={Code as Reward: Empowering Reinforcement Learning with VLMs},
  author={Venuto, David and Islam, Mohammad Sami Nur and Klissarov, Martin and Precup, Doina and Yang, Sherry and Anand, Ankit},
  booktitle={Proceedings of the 41st International Conference on Machine Learning},
  year={2024}
}

@inproceedings{guan2024tasksuccess,
  title={{Task Success} is not Enough: Investigating the Use of Video-Language Models as Behavior Critics for Catching Undesirable Agent Behaviors},
  author={Guan, Lin and Zhou, Yifan and Liu, Denis and Zha, Yantian and Ben Amor, Heni and Kambhampati, Subbarao},
  booktitle={First Conference on Language Modeling},
  year={2024},
  url={https://openreview.net/forum?id=otKo4zFKmH}
}

@inproceedings{NEURIPS2023_d9042abf,
  author={Escontrela, Alejandro and Adeniji, Ademi and Yan, Wilson and Jain, Ajay and Peng, Xue Bin and Goldberg, Ken and Lee, Youngwoon and Hafner, Danijar and Abbeel, Pieter},
  booktitle={Advances in Neural Information Processing Systems},
  pages={68760--68783},
  publisher={Curran Associates, Inc.},
  title={Video Prediction Models as Rewards for Reinforcement Learning},
  url={https://proceedings.neurips.cc/paper_files/paper/2023/file/d9042abf40782fbce28901c1c9c0e8d8-Paper-Conference.pdf},
  volume={36},
  year={2023},
}

@article{baumli2023visionlanguage,
  title={Vision-Language Models as a Source of Rewards},
  author={Baumli, Kate and Baveja, Satinder and Behbahani, Feryal and Chan, Harris and Comanici, Gheorghe and Flennerhag, Sebastian and Gazeau, Maxime and Holsheimer, Kristian and Horgan, Dan and Laskin, Michael and Lyle, Clare and Masoom, Hussain and McKinney, Kay and Mnih, Volodymyr and Neitz, Alexander and Nikulin, Dmitry and Pardo, Fabio and Parker-Holder, Jack and Quan, John and Rockt{\"a}schel, Tim and Sahni, Himanshu and Schaul, Tom and Schroecker, Yannick and Spencer, Stephen and Steigerwald, Richie and Wang, Luyu and Zhang, Lei},
  journal={arXiv preprint arXiv:2312.09187},
  year={2023}
}

@inproceedings{rocamonde2024visionlanguage,
  title={Vision-Language Models are Zero-Shot Reward Models for Reinforcement Learning},
  author={Rocamonde, Juan and Montesinos, Victoriano and Nava, Elvis and Perez, Ethan and Lindner, David},
  booktitle={The Twelfth International Conference on Learning Representations},
  year={2024},
  url={https://openreview.net/forum?id=N0I2RtD8je}
}

@inproceedings{ma2024eureka,
  title={Eureka: Human-Level Reward Design via Coding Large Language Models},
  author={Ma, Yecheng Jason and Liang, William and Wang, Guanzhi and Huang, De-An and Bastani, Osbert and Jayaraman, Dinesh and Zhu, Yuke and Fan, Linxi and Anandkumar, Anima},
  booktitle={The Twelfth International Conference on Learning Representations},
  year={2024},
  url={https://openreview.net/forum?id=IEduRUO55F}
}

@misc{yang2023robotfinetuningeasypretraining,
  title={Robot Fine-Tuning Made Easy: Pre-Training Rewards and Policies for Autonomous Real-World Reinforcement Learning},
  author={Jingyun Yang and Max Sobol Mark and Brandon Vu and Archit Sharma and Jeannette Bohg and Chelsea Finn},
  year={2023},
  eprint={2310.15145},
  archivePrefix={arXiv},
  primaryClass={cs.RO},
  url={https://arxiv.org/abs/2310.15145},
}

@inproceedings{xietext2reward,
  title={Text2Reward: Reward Shaping with Language Models for Reinforcement Learning},
  author={Xie, Tianbao and Zhao, Siheng and Wu, Chen Henry and Liu, Yitao and Luo, Qian and Zhong, Victor and Yang, Yanchao and Yu, Tao},
  booktitle={The Twelfth International Conference on Learning Representations},
  year={2024},
  url={https://openreview.net/forum?id=tUM39YTRxH}
}

@misc{bhateja2023roboticofflinerlinternet,
  title={Robotic Offline RL from Internet Videos via Value-Function Pre-Training},
  author={Chethan Bhateja and Derek Guo and Dibya Ghosh and Anikait Singh and Manan Tomar and Quan Vuong and Yevgen Chebotar and Sergey Levine and Aviral Kumar},
  year={2023},
  eprint={2309.13041},
  archivePrefix={arXiv},
  primaryClass={cs.RO},
  url={https://arxiv.org/abs/2309.13041},
}

@inproceedings{pmlr-v202-ma23b,
  title={{LIV}: Language-Image Representations and Rewards for Robotic Control},
  author={Ma, Yecheng Jason and Kumar, Vikash and Zhang, Amy and Bastani, Osbert and Jayaraman, Dinesh},
  booktitle={Proceedings of the 40th International Conference on Machine Learning},
  pages={23301--23320},
  year={2023},
  volume={202},
  series={Proceedings of Machine Learning Research},
  publisher={PMLR},
  url={https://proceedings.mlr.press/v202/ma23b.html},
}

@inproceedings{pmlr-v205-nair23a,
  title={{R3M}: A Universal Visual Representation for Robot Manipulation},
  author={Nair, Suraj and Rajeswaran, Aravind and Kumar, Vikash and Finn, Chelsea and Gupta, Abhinav},
  booktitle={Proceedings of The 6th Conference on Robot Learning},
  pages={892--909},
  year={2023},
  editor={Liu, Karen and Kulic, Dana and Ichnowski, Jeff},
  volume={205},
  series={Proceedings of Machine Learning Research},
  publisher={PMLR},
  url={https://proceedings.mlr.press/v205/nair23a.html},
}

@inproceedings{pmlr-v229-yu23a,
  title={Language to Rewards for Robotic Skill Synthesis},
  author={Yu, Wenhao and Gileadi, Nimrod and Fu, Chuyuan and Kirmani, Sean and Lee, Kuang-Huei and Gonzalez Arenas, Montserrat and Chiang, Hao-Tien Lewis and Erez, Tom and Hasenclever, Leonard and Humplik, Jan and Ichter, Brian and Xiao, Ted and Xu, Peng and Zeng, Andy and Zhang, Tingnan and Heess, Nicolas and Sadigh, Dorsa and Tan, Jie and Tassa, Yuval and Xia, Fei},
  booktitle={Proceedings of The 7th Conference on Robot Learning},
  pages={374--404},
  year={2023},
  editor={Tan, Jie and Toussaint, Marc and Darvish, Kourosh},
  volume={229},
  series={Proceedings of Machine Learning Research},
  publisher={PMLR},
  url={https://proceedings.mlr.press/v229/yu23a.html}
}

@inproceedings{pmlr-v235-liu24o,
  title={{PEARL}: Zero-shot Cross-task Preference Alignment and Robust Reward Learning for Robotic Manipulation},
  author={Liu, Runze and Du, Yali and Bai, Fengshuo and Lyu, Jiafei and Li, Xiu},
  booktitle={Proceedings of the 41st International Conference on Machine Learning},
  pages={30946--30964},
  year={2024},
  volume={235},
  series={Proceedings of Machine Learning Research},
  publisher={PMLR},
  url={https://proceedings.mlr.press/v235/liu24o.html}
}

@inproceedings{pmlr-v232-du23b,
  title={Vision-Language Models as Success Detectors},
  author={Du, Yuqing and Konyushkova, Ksenia and Denil, Misha and Raju, Akhil and Landon, Jessica and Hill, Felix and de Freitas, Nando and Cabi, Serkan},
  booktitle={Proceedings of The 2nd Conference on Lifelong Learning Agents},
  pages={120--136},
  year={2023},
  editor={Chandar, Sarath and Pascanu, Razvan and Sedghi, Hanie and Precup, Doina},
  volume={232},
  series={Proceedings of Machine Learning Research},
  publisher={PMLR},
  url={https://proceedings.mlr.press/v232/du23b.html}
}

@misc{ma2023vipuniversalvisualreward,
  title={VIP: Towards Universal Visual Reward and Representation via Value-Implicit Pre-Training},
  author={Yecheng Jason Ma and Shagun Sodhani and Dinesh Jayaraman and Osbert Bastani and Vikash Kumar and Amy Zhang},
  year={2023},
  eprint={2210.00030},
  archivePrefix={arXiv},
  primaryClass={cs.RO},
  url={https://arxiv.org/abs/2210.00030},
}

@inproceedings{pmlr-v162-mahmoudieh22a,
  title={Zero-Shot Reward Specification via Grounded Natural Language},
  author={Mahmoudieh, Parsa and Pathak, Deepak and Darrell, Trevor},
  booktitle={Proceedings of the 39th International Conference on Machine Learning},
  pages={14743--14752},
  year={2022},
  editor={Chaudhuri, Kamalika and Jegelka, Stefanie and Song, Le and Szepesvari, Csaba and Niu, Gang and Sabato, Sivan},
  volume={162},
  series={Proceedings of Machine Learning Research},
  month={17--23 Jul},
  publisher={PMLR},
  url={https://proceedings.mlr.press/v162/mahmoudieh22a.html}
}

@inproceedings{pmlr-v168-cui22a,
  title={Can Foundation Models Perform Zero-Shot Task Specification For Robot Manipulation?},
  author={Cui, Yuchen and Niekum, Scott and Gupta, Abhinav and Kumar, Vikash and Rajeswaran, Aravind},
  booktitle={Proceedings of The 4th Annual Learning for Dynamics and Control Conference},
  pages={893--905},
  year={2022},
  volume={168},
  series={Proceedings of Machine Learning Research},
  publisher={PMLR},
  url={https://proceedings.mlr.press/v168/cui22a.html}
}

@inproceedings{NEURIPS2021_868b7df9,
  author={Lee, Youngwoon and Szot, Andrew and Sun, Shao-Hua and Lim, Joseph J},
  booktitle={Advances in Neural Information Processing Systems},
  pages={16118--16130},
  publisher={Curran Associates, Inc.},
  title={Generalizable Imitation Learning from Observation via Inferring Goal Proximity},
  url={https://proceedings.neurips.cc/paper_files/paper/2021/file/868b7df964b1af24c8c0a9e43a330c6a-Paper.pdf},
  volume={34},
  year={2021},
}

@inproceedings{pmlr-v164-nair22a,
  title={Learning Language-Conditioned Robot Behavior from Offline Data and Crowd-Sourced Annotation},
  author={Nair, Suraj and Mitchell, Eric and Chen, Kevin and ichter, brian and Savarese, Silvio and Finn, Chelsea},
  booktitle={Proceedings of the 5th Conference on Robot Learning},
  pages={1303--1315},
  year={2022},
  volume={164},
  series={Proceedings of Machine Learning Research},
  publisher={PMLR},
  url={https://proceedings.mlr.press/v164/nair22a.html}
}

@misc{chen2021learninggeneralizableroboticreward,
  title={Learning Generalizable Robotic Reward Functions from "In-The-Wild" Human Videos},
  author={Annie S. Chen and Suraj Nair and Chelsea Finn},
  year={2021},
  eprint={2103.16817},
  archivePrefix={arXiv},
  primaryClass={cs.RO},
  url={https://arxiv.org/abs/2103.16817},
}

@misc{cabi2020scalingdatadrivenroboticsreward,
  title={Scaling data-driven robotics with reward sketching and batch reinforcement learning},
  author={Serkan Cabi and Sergio G{\'o}mez Colmenarejo and Alexander Novikov and Ksenia Konyushkova and Scott Reed and Rae Jeong and Konrad Zolna and Yusuf Aytar and David Budden and Mel Vecerik and Oleg Sushkov and David Barker and Jonathan Scholz and Misha Denil and Nando de Freitas and Ziyu Wang},
  year={2020},
  eprint={1909.12200},
  archivePrefix={arXiv},
  primaryClass={cs.RO},
  url={https://arxiv.org/abs/1909.12200},
}

@misc{singh2019endtoendroboticreinforcementlearning,
  title={End-to-End Robotic Reinforcement Learning without Reward Engineering},
  author={Avi Singh and Larry Yang and Kristian Hartikainen and Chelsea Finn and Sergey Levine},
  year={2019},
  eprint={1904.07854},
  archivePrefix={arXiv},
  primaryClass={cs.LG},
  url={https://arxiv.org/abs/1904.07854},
}

@misc{sermanet2017unsupervisedperceptualrewardsimitation,
  title={Unsupervised Perceptual Rewards for Imitation Learning},
  author={Pierre Sermanet and Kelvin Xu and Sergey Levine},
  year={2017},
  eprint={1612.06699},
  archivePrefix={arXiv},
  primaryClass={cs.CV},
  url={https://arxiv.org/abs/1612.06699},
}

@misc{yan2026progressvlaprogressguideddiffusionpolicy,
  title={ProgressVLA: Progress-Guided Diffusion Policy for Vision-Language Robotic Manipulation},
  author={Hongyu Yan and Qiwei Li and Jiaolong Yang and Yadong Mu},
  year={2026},
  eprint={2603.27670},
  archivePrefix={arXiv},
  primaryClass={cs.RO},
  url={https://arxiv.org/abs/2603.27670},
}

@misc{zhang2026recurrentreasoningvisionlanguagemodels,
  title={Recurrent Reasoning with Vision-Language Models for Estimating Long-Horizon Embodied Task Progress},
  author={Yuelin Zhang and Sijie Cheng and Chen Li and Zongzhao Li and Yuxin Huang and Yang Liu and Wenbing Huang},
  year={2026},
  eprint={2603.17312},
  archivePrefix={arXiv},
  primaryClass={cs.CV},
  url={https://arxiv.org/abs/2603.17312},
}

@inproceedings{garmentlab,
 author = {Lu, Haoran and Wu, Ruihai and Li, Yitong and Li, Sijie and Zhu, Ziyu and Ning, Chuanruo and Shen, Yan and Luo, Longzan and Chen, Yuanpei and Dong, Hao},
 booktitle = {Advances in Neural Information Processing Systems},
 doi = {10.52202/079017-0379},
 editor = {A. Globerson and L. Mackey and D. Belgrave and A. Fan and U. Paquet and J. Tomczak and C. Zhang},
 pages = {11866--11903},
 publisher = {Curran Associates, Inc.},
 title = {GarmentLab: A Unified Simulation and Benchmark for Garment Manipulation},
 url = {https://proceedings.neurips.cc/paper_files/paper/2024/file/15f80ec0fed53885d2ca6272edb96ede-Paper-Conference.pdf},
 volume = {37},
 year = {2024}
}

@INPROCEEDINGS{11128651,
  author={Wu, Ruihai and Chen, Haozhe and Zhang, Mingtong and Lu, Haoran and Li, Yitong and Li, Yunzhu},
  booktitle={2025 IEEE International Conference on Robotics and Automation (ICRA)}, 
  title={Neural Dynamics Augmented Diffusion Policy}, 
  year={2025},
  volume={},
  number={},
  pages={13234-13241},
  doi={10.1109/ICRA55743.2025.11128651}}

@misc{Intelligence2026pi07AS,
  author = {{Physical Intelligence}},
  title = {Pi-0.7: A Steerable Generalist Robotic Foundation Model with Emergent Capabilities},
  year = {2026},

  howpublished = {arXiv preprint},
  note = {CorpusID: 287607456}
}

@misc{generalist2025gen0,
  author = {{Generalist AI Team}},
  title = {GEN-0: Embodied Foundation Models That Scale with Physical Interaction},
  year = {2025},
  howpublished = {Generalist AI Blog},
  note = {November 4, 2025}
}

@misc{Nvidia2025GR00TNA,
  author = {{NVIDIA}},
  title = {GR00T N1: An Open Foundation Model for Generalist Humanoid Robots},
  year = {2025},
  howpublished = {arXiv:2503.14734}
}

@article{Intelligence202505AV,
  title={$\pi$0.5: a Vision-Language-Action Model with Open-World Generalization},
  author={Physical Intelligence and Kevin Black and Noah Brown and James Darpinian and Karan Dhabalia and Danny Driess and Adnan Esmail and Michael Equi and Chelsea Finn and Niccolo Fusai and Manuel Y. Galliker and Dibya Ghosh and Lachy Groom and Karol Hausman and Brian Ichter and Szymon Jakubczak and Tim Jones and Liyiming Ke and Devin LeBlanc and Sergey Levine and Adrian Li-Bell and Mohith Mothukuri and Suraj Nair and Karl Pertsch and Allen Z. Ren and Lucy Xiaoyang Shi and Laura Smith and Jost Tobias Springenberg and Kyle Stachowicz and James Tanner and Quan Vuong and Homer Rich Walke and Anna Walling and Haohuan Wang and Lili Yu and Ury Zhilinsky},
  journal={ArXiv},
  year={2025},
  volume={abs/2504.16054},
  url={https://api.semanticscholar.org/CorpusID:277993634}
}

@misc{yuan2026fastwamworldactionmodels,
      title={Fast-WAM: Do World Action Models Need Test-time Future Imagination?}, 
      author={Tianyuan Yuan and Zibin Dong and Yicheng Liu and Hang Zhao},
      year={2026},
      eprint={2603.16666},
      archivePrefix={arXiv},
      primaryClass={cs.CV},
      url={https://arxiv.org/abs/2603.16666}, 
}

@misc{Ye2026WorldAM,
  author = {Ye, Seonghyeon and Ge, Yunhao and Zheng, Kaiyuan and Jang, Joel},
  title = {World Action Models are Zero-shot Policies},
  year = {2026},
  howpublished = {arXiv:2602.15922}
}

@misc{liu2023liberobenchmarkingknowledgetransfer,
      title={LIBERO: Benchmarking Knowledge Transfer for Lifelong Robot Learning}, 
      author={Bo Liu and Yifeng Zhu and Chongkai Gao and Yihao Feng and Qiang Liu and Yuke Zhu and Peter Stone},
      year={2023},
      eprint={2306.03310},
      archivePrefix={arXiv},
      primaryClass={cs.AI},
      url={https://arxiv.org/abs/2306.03310}, 
}

@article{chen2026robodojo,
  title={RoboDojo: A Unified Sim-and-Real Benchmark for Comprehensive Evaluation of Generalist Robot Manipulation Policies},
  author={Chen, Tianxing and Chen, Yue and Li, Zixuan and Tang, Junyuan and Su, Kailun and Wan, Weijie and Chen, Baijun and Lu, Haoran and Yan, Haowen and Su, Honghao and others},
  journal={arXiv preprint arXiv:2607.04434},
  year={2026}
}

@misc{lu2026magicsimunifiedinfrastructureexecutable,
      title={MagicSim: A Unified Infrastructure for Executable Embodied Interaction},
      author={Haoran Lu and Songling Liu and Yue Chen and Guo Ye and Mutian Shen and Shuyang Yu and Yu Xiao and Jihai Zhao and Shang Wu and Jianshu Zhang and Xiangtian Gui and Chuye Hong and Yuran Wang and Maojiang Su and Jiayi Wang and Ruihai Wu and Zhaoran Wang and Han Liu},
      year={2026},
      eprint={2606.17511},
      archivePrefix={arXiv},
      primaryClass={cs.RO},
      url={https://arxiv.org/abs/2606.17511},
}

@misc{chen2026rmbenchmemorydependentroboticmanipulation,
      title={RMBench: Memory-Dependent Robotic Manipulation Benchmark with Insights into Policy Design}, 
      author={Tianxing Chen and Yuran Wang and Mingleyang Li and Yan Qin and Hao Shi and Zixuan Li and Yifan Hu and Yingsheng Zhang and Kaixuan Wang and Yue Chen and Hongcheng Wang and Renjing Xu and Ruihai Wu and Yao Mu and Yaodong Yang and Hao Dong and Ping Luo},
      year={2026},
      eprint={2603.01229},
      archivePrefix={arXiv},
      primaryClass={cs.RO},
      url={https://arxiv.org/abs/2603.01229}, 
}

\end{document}